\documentclass[sigconf]{acmart}

\usepackage{bbding}
\usepackage{multirow}
\usepackage{array}
\usepackage{arydshln}
\newcolumntype{H}{>{\setbox0=\hbox\bgroup}c<{\egroup}@{}}
\AtBeginDocument{%
  \providecommand\BibTeX{{%
    \normalfont B\kern-0.5em{\scshape i\kern-0.25em b}\kern-0.8em\TeX}}}
    
    \renewcommand\footnotetextcopyrightpermission[1]{}
\setcopyright{acmcopyright}
\copyrightyear{2022}
\acmYear{2022}
\setcopyright{acmcopyright}

\acmConference[WSDM '22] {Proceedings of the Fifteenth ACM International Conference on Web Search and Data Mining}{February 21--25, 2022}{Tempe, AZ, USA.}
\acmBooktitle{Proceedings of the Fifteenth ACM International Conference on Web Search and Data Mining (WSDM '22), February 21--25, 2022, Tempe, AZ, USA}
\acmPrice{15.00}
\acmISBN{978-1-4503-9132-0/22/02}

\newcommand{\name}{\textsf{SPARTA}}
\newcommand{\dataset}{\textsf{HOPE}}
\usepackage{soul}
\usepackage{appendix}
\usepackage{enumitem}

\begin{document}
\fancyhead{}
\title{Speaker and Time-aware Joint Contextual Learning for Dialogue-act Classification in Counselling Conversations}

 \author{Ganeshan Malhotra$^1$, Abdul Waheed$^2$, Aseem Srivastava$^3$}
 \author{Md Shad Akhtar$^3$, Tanmoy Chakraborty$^3$}
 \affiliation{$^1$BITS Pilani, Goa, India, $^2$Maharaja Agrasen Institute of Technology, New Delhi, India, $^3$IIIT-Delhi, India}
 \email{{ganeshanmalhotra, abdulwaheed1513}@gmail.com, {aseems,shad.akhtar,tanmoy}@iiitd.ac.in}

\begin{abstract}
The onset of the COVID-19 pandemic has brought the mental health of people under risk. Social counselling has gained remarkable significance in this environment. Unlike general goal-oriented dialogues, a conversation between a patient and a therapist is considerably implicit, though the objective of the conversation is quite apparent. In such a case, understanding the intent of the patient is imperative in providing effective counselling in therapy sessions, and the same applies to a dialogue system as well. 
In this work, we take forward a small but an important step in the development of an automated dialogue system for mental-health counselling. We develop a novel dataset, named \dataset, to provide a platform for the dialogue-act classification in counselling conversations. We identify the requirement of such conversation and propose twelve domain-specific dialogue-act (DAC) labels. We collect $\sim 12.9K$ utterances from publicly-available counselling session videos on YouTube, extract their transcripts, clean, and annotate them with DAC labels. Further, we propose \name, a transformer-based architecture with a novel speaker- and time-aware contextual learning for the dialogue-act classification. Our evaluation shows convincing performance over several baselines, achieving state-of-the-art on \dataset. We also supplement our experiments with extensive empirical and qualitative analyses of \name. 

\end{abstract}




\maketitle
\section{Introduction}
Mental illness remains an alarming  global health issue today. Due to the COVID-19 pandemic, there has been a significant growth in mental health disorders such as depression, attention deficit hyperactivity disorder (ADHD) and hypertension \cite{Moreno2020}. 
A recent study shows an unprecedented $20$\% increase in patients with mental health illness\footnote{\url{https://cutt.ly/WbEziBF/}}. Similar study discusses the adverse impact on the mental health of US college students due to the pandemic \cite{wang2020}.

\begin{table}[t]
    \caption{Example of a sample conversation session between a patient and a therapist. Each utterance has an associated dialogue-act classification (DAC) label.}
    \label{tab:TableOfUtterances}
    \vspace{-3mm}
    \resizebox{0.98\columnwidth}{!}{
    \begin{tabular}{p{0.78\linewidth} p{0.2\linewidth}}
        \toprule[1pt]
        \multicolumn{1}{l}{\textbf{Utterance}} & \textbf{DAC labels}\\
        \toprule[1pt]
        \textcolor{blue}{Therapist:} \textit{Jackie, how are you?} & Greeting \\  \cline{1-2}

        \textcolor{red}{Patient:} \textit{Okay, How are you?} & Greeting \\ \cline{1-2}

        \textcolor{blue}{Therapist:} \textit{Thanks for asking. I see that you have signed a release so I could talk to your mother and that she brought you in today. What's going on there?} & Information Request\\
        \cline{1-2}

        \textcolor{red}{Patient:} \textit{They think I have a drinking problem. My family..} & Information Delivery\\
        \cline{1-2}

        \textcolor{blue}{Therapist:} \textit{Your family thinks you have a drinking problem?} & Clarification Request\\
        \cline{1-2}

        \textcolor{red}{Patient:} \textit{Yeah. So we really started this was this past weekend. They came to pick me up for my bridal shower. And I was I was drunk when they came to get me so I couldn't go and now everybody's pretty pissed at me.} & Clarification Delivery \\
        \cline{1-2}

        \textcolor{blue}{Therapist:} \textit{So they asked you to come into the agency?} & Clarification Request\\
        \cline{1-2}

        \textcolor{red}{Patient:} \textit{Yeah, you know, I don't want them to hate me or anything. So I agreed to come.} & Clarification Delivery\\ \toprule[1pt]
    \end{tabular}}
    \vspace{-5mm}
\end{table}

Counselling therapy can benefit many people under risk by providing them emotional support. Amidst the surge in the number of patients, it has become  a challenge for the therapists to diagnose too many patients. On the other hand, patients have found it difficult to access the services of the therapist amid lockdown.

Counseling therapy is a sophisticated procedure that deals with the expression of emotion and intent of patients with different personalities. 
To build a strong therapeutic relationship with the patient, it is essential for a therapist to develop a better understanding of the implicit intents of the patients.
The nature of conversations in a social counselling setting is particularly distinct as compared to a conventional chit-chat or goal-oriented conversations.
It follows a pattern which is different from both goal-oriented and general chit-chat based conversations. Usually these conversations begin with greetings followed by the therapist inquiring for problems faced by the patient. The therapist usually delves deeper into a particular problem acquiring as much context and fine-grained information before advising a remedy. These conversations also heavily utilise the contextual information of the entire conversation history. Moreover, the prime objective of the conversation is to understand the explicit and implicit requirements of the patients and suggest potential solutions accordingly. In comparison, a traditional goal-oriented dialogue system does not regard any implicit requirements, whereas a chit-chat based system lacks a target and does not care about the final solution. Another major difference is the length of utterances and conversations in a counseling session. These are particularly lengthy as patients describe their difficulties and issues, while the therapists list out possible causes and preventive solutions. 

The task of Dialogue-act classification (DAC) is cardinal in a  dialogue system and even more so in counselling based conversations. It deals with understanding the intended requirements of the utterances, which essentially act as one of the precursors for the dialogue response generation. 
For instance, we present an example of a therapy session in Table \ref{tab:TableOfUtterances}. For each utterance in Table \ref{tab:TableOfUtterances}, a corresponding label defines its dialogue-act. As we can observe from the first two utterances, they are a part of the complementary greetings that usually occur at the beginning of a natural conversation. Subsequently, in the third utterance, the therapist leads the conversation and requests for information. In response, the patient delivers the requested information.

Earlier studies like \cite{ahmadvand2019contextual,raheja-tetreault-2019-dialogue} tackle the task of dialogue-act classification on chit-chat based conversation datasets such as Switchboard corpus \cite{225858}. Their proposed architectures take into account the contextual dependency of an utterance that aids in efficient dialogue-act classification. For example, an utterance tagged as having a dialogue-act `question' has a high probability of being followed by an utterance with tag `answer'. In another work, \citet{shang-etal-2020-speaker} argued that the information of speaker change is a critical feature in the dialogue-act classification task.

Considering the severity of the issue and the complexity of the task, designing an automated system can facilitate the counselling sessions or assist the therapist, thus allowing them to cater to more patients. 
Literature in the natural language processing domain suggests a significant effort in understanding and building models for conversational dialogue \cite{li2017dailydialog, cerisara2018multi}. However, there are hardly any models that support mental-health counseling as a dialogue system; this is primarily due to lack of data. In this paper, we aim to address these limitations by creating the \dataset\footnote{Mental {\bf H}ealth c{\bf O}unselling of {\bf P}ati{\bf E}nts} dataset which consists of therapy conversations covering cognitive-behavioral therapy (CBT), child therapy, family therapy, etc. 
The \dataset\ dataset contains $\sim 12.9K$ utterances across $212$ mental-health counseling sessions. Each utterance in the dataset is tagged with one of the $12$ counseling-aligned dialogue-act labels (c.f. Section \ref{sec:dataset}).

We also propose \name\footnote{{\bf SP}eaker and time-{\bf A}wa{\bf R}e con{\bf T}extual tr{\bf A}nsformer}, a novel speaker- and time-aware contextual transformer model for dialogue-act classification. \name\ exploits both the local and global contexts along with the speaker-dynamics in the dialogue. We model the problem as a dialogue-level sequence classification task, where the aim is to predict an appropriate dialogue-act for each utterance in a dialogue. 
To incorporate the global context, we employ a Gated Recurrent Unit (GRU) \cite{cho-etal-2014-learning} that takes an utterance representation at each step of the dialogue. In addition, we introduce a novel time-aware attention mechanism to capture the local context -- a sliding-window based memory unit is maintained, and subsequently, a cross-attention between the current utterance and the memory unit is computed.
Our evaluation shows substantial improvement in performance in comparison to the recent state-of-the-art systems. Furthermore, we provide empirical evidences for each module of \name\ using an extensive ablation study and detailed analyses.

\subsubsection*{\bf Major contributions:} We summarize the main contributions of our current work as follows:
\begin{itemize}[leftmargin=*]
\setlength{\itemsep}{0pt}
    \item We present \dataset, a novel and large-scale manually annotated, counselling-based conversation data for dialogue-act classification. To the best of our knowledge, the current study is one of the first efforts in compiling a dataset related to the mental-health counseling dialogue system. 
    \item To cater to the requirements of counseling conversations, we define a novel hierarchical annotation scheme for the dialogue-act annotation. We propose twelve dialogue-act labels that are aligned with mental-health counseling session.
    \item We propose \name, a novel dialogue-act classification system that combines speaker-dynamics and local context through a time-aware attention mechanism, along with long-term global context.
    \item We perform extensive ablation study to establish the efficacy of each module of \name. Furthermore, the comparative analysis shows that it attains state-of-the-art performance on \dataset. 
\end{itemize}

\subsubsection*{\bf{Societal Impact: }}
A significant increase in the number of mental health issues has been observed in the last few years. The lack of therapists is a stumbling block to the mental health of society.  Therapist-Bots (mental health chatbots) could bridge the gap by effectively interacting with patients and understanding them. Conversely, end-to-end chatbots in the mental health domain are delicate, where every aspect of the therapy is needed to be perceived precisely. Our research aims at the dialogue-understanding module in the mental health conversational system. The ongoing research in mental health domain could exploit this work and benefit the chatbots to understand the therapy conversation in a better way.

\subsubsection*{\bf Reproducibility:} 
We have made the code and (a subset of) the dataset available at github\footnote{\url{https://github.com/LCS2-IIITD/SPARTA_WSDM2022}}.

\section{Related Work}
The current work is connected to the existing literature in at least two dimensions -- first, the dialogue-act classification, and second, text processing for mental-health counseling. We present our literature survey for both dimensions. 

\textbf{Mental Health and the role of text processing.}
The impact of Natural language processing in the study of mental-health is substantial. 
Though the field of therapeutic discourse analysis has been around since  1960s \cite{van2011discourse}, research on dialogue systems in mental-health domain is in nascent stage. Previous research on mental health intervention systems primarily focused on the problems related to the suicide ideation prevention \cite{mishra2019snap} or generating empathetic responses to users \cite{naous2020empathy}. \citet{levis:leonard:westgate:gui:watts:shiner:2020} studied the mental-health notes to detect suicide ideation, whereas, \citet{fine2020assessing} employed text processing to detect symptoms of anxiety and depression using social media text. In another work, \citet{Pennebaker2003PsychologicalAO} showed the importance of several keywords in revealing users' social and psychological behaviours. \citet{10.1145/3357384.3357937} proposed an emotion-aware model for human-like emotional conversations.

\citet{gratch2014distress} presented the Distress Analysis Interview Corpus (DAIC) to identify verbal and non-verbal cues of distress during an interview. Among other methods, the data collection was done using the automated agent Ellie based on the work of \cite{morbini_demonstration_2014}.
Recently, \citet{pmid30858710} presented a survey of the chat-bots in the mental-health domain. The authors compared the strengths and weaknesses of three existing conversational agents, namely Wysa\footnote{https://www.wysa.io/}, Woebot\footnote{https://woebothealth.com/}, and Joy\footnote{ https://www.hellojoy.ai/support}. The drawbacks of these systems are that some of these are rule-based, while others are primarily data collection module for an offline counselling. In comparison, ours is the effort in the development of an online counselling system.

\textbf{Dialogue-act Classification.}
Studies on dialogue systems have always fascinated researchers ever since ELIZA \cite{eliza:1966}, the first rule-based system  was developed. 
The dialogue-act classification module is one of the most critical components of a dialogue agent which caters to the requirement of the dialogue system by serving at the natural language understanding helm of the dialogue system. 

Previous research treats the problem of dialogue-act classification either as a standalone text classification task or a sequence labelling task. Recently, \citet{Colombo2020GuidingAI} suggested a sequence-to-sequence text generation approach for the dialogue-act classification.
Earlier studies like \citet{reithinger1997dialogue} and \citet{grau2004dialogue} focused on lexical, syntactical, and prosodic features for classification. 
In another work, \citet{ortega2019context} employed CNNs \cite{10.1162/neco.1989.1.4.541} and CRFs \cite{10.5555/645530.655813}.
\citet{lee-dernoncourt-2016-sequential} proposed a method based on CNNs and RNNs \cite{10.5555/104279.104293} that used the previous contextual utterances to predict the dialogue-act of the current utterance.

 \citet{ahmadvand2019contextual} proposed a contextual dialogue-act classifier (CDAC) and used transfer-learning to train their model on human-machine conversations. The model proposed by \citet{raheja2019dialogue} uses self-attention mechanism on RNNs to achieve impressive results on benchmark datasets. However, these works do not take into account the speaker-level information which is imperative in social-counseling based conversations. \citet{10.1145/3340531.3411967} proposed a method to detect relevant context in retrieval-based dialogue systems. \citet{chen2018dialogue} proposed a CRF-attentive structured network to capture the long-range contextual dependencies using structured attention mechanism. \citet{10.1145/3357384.3358145} proposed to classify concurrent dialogue-acts of an utterance by modelling the contextual features.
Recently, \citet{DBLP:conf/aaai/QinCLN020} used co-interactive relation networks to jointly capture sentiment and the associated dialogue-acts with an utterance. Their model achieved significant results on Mastodon \cite{cerisara2018multi} and DailyDialog \cite{li2017dailydialog} datasets. Similarly, \citet{Saha2020EmotionAD} jointly learns the dialogue-act classification and emotion recognition tasks in a multi-modal setup.

To the best of our knowledge,  the model by \citet{Shang2020SpeakerchangeAC} is the first one which takes speaker transitions for DAC into account. It uses a modified version of CRFs to capture the speaker-change and achieves state-of-the-art results on SwitchBoard dataset \cite{225858}. Similar to earlier works, we also treat DAC as a dialogue-level sequence labelling task. We jointly take the global and local contexts of the conversation and the speaker of the utterance as the key factors for the classification. We hypothesize that such information offers crucial clues at different stages of the model. In contrast, the existing systems had incorporated the role of global context and speakers dynamic independently. 
\vspace{-0.22cm}
\section{Dataset}
\label{sec:dataset}
In this section, we present our dialogue-act classification dataset, called \dataset. In total, we annotated $\sim 12.9K$ utterances with $12$ dialogue-act labels carefully designed to cater to the requirements of a counseling session. The remaining section furnishes the details of data collection, annotation schemes, dialogue-act labels and necessary statistics.

\vspace{-0.2cm}
\subsubsection*{\bf Data Collection}
One of major hurdles we faced in the process of data collection was the unavailability of public counseling sessions, mainly due to the fact that they usually contain sensitive personal information. To curate this data, we carefully explored the web and collected publicly-available pre-recorded counselling videos on YouTube.
To ensure confidentiality, we randomly assign synthetic names to all patients and therapists in all examples.

In the next step, we extract the transcriptions of each video using OTTER (https://www.otter.ai/), an automatic speech recognition tool. Subsequently, we correct transcription errors to remove any noise (i.e., spelling or grammatical mistakes). The data collection process provides us $12.9K$ utterances from 212 counseling therapy sessions -- all of them are dyadic conversations only.

\begin{table*}[t]
    \caption{Dialogue-act distribution in \dataset. The train, test, and validation splits are 70:20:10.}
    \centering
    \resizebox{0.7\textwidth}{!}
    {
    \begin{tabular}{c H c c c c c c c c c c c c c c}
    \toprule[1pt]
         &  &  & \multicolumn{11}{c}{\bf Dialogue-act labels} & \\ \cline{4-15}
        \textbf{Split} & \textbf{\#Dialogues} & \textbf{Speaker} & ID & IRQ & GT & GC & CRQ & YNQ & CD & ACK & PA & NA & OD & ORQ & \textbf{Total} \\
        \toprule[1pt]
        \multirow{2}{*}{\textbf{Train}} & \multirow{2}{*}{149} & \textbf{Patient} & 1886 & 23 & 155 & 494 & 77 & 12 & 666 & 607 & 269 & 271 & 205 & 3 & 4668\\
         & & \textbf{Therapist} & 393 & 1474 & 212 & 829 & 710 & 692 & 83 & 125 & 6 & 7 & 40 & 180 & 4751\\
        \cdashline{1-1} \cdashline{3-16} 
        \multirow{2}{*}{\textbf{Test}} & \multirow{2}{*}{42} & \textbf{Patient} & 466 & 7 & 32 & 122 & 13 & 6 & 189 & 97 & 73 & 66 & 48 & - & 1119\\
         & & \textbf{Therapist} & 66 & 376 & 44 & 209 & 199 & 134 & 15 & 31 & 6 & - & 8 & 34 & 1122\\
        \cdashline{1-1} \cdashline{3-16}
        \multirow{2}{*}{\textbf{Val}} & \multirow{2}{*}{21} & \textbf{Patient} & 246 & 3 & 17 & 83 & 8 & - & 97 & 63 & 34 & 27 & 16 & 1 & 595\\
         & & \textbf{Therapist} & 41 & 194 & 22 & 113 & 101 & 85 & 10 & 16 & - & - & 3 & 14 & 599\\  \cline{1-16}
         \multirow{2}{*}{\textbf{Total}} & \multirow{2}{*}{212} & \textbf{Patient} & 2598 & 33 & 204 & 699 & 98 & 18 & 952 & 767 & 376 & 364 & 269 & 4 & 6382\\
         & & \textbf{Therapist} & 500 & 2044 & 278 & 1151 & 1010 & 911 & 108 & 172 & 12 & 7 & 51 & 228 & 6472\\ 
         \toprule[1pt]

    \end{tabular}}
    \label{tab:counts}
\end{table*}

\begin{figure}[t]
  \centering
  \includegraphics[width=\columnwidth]{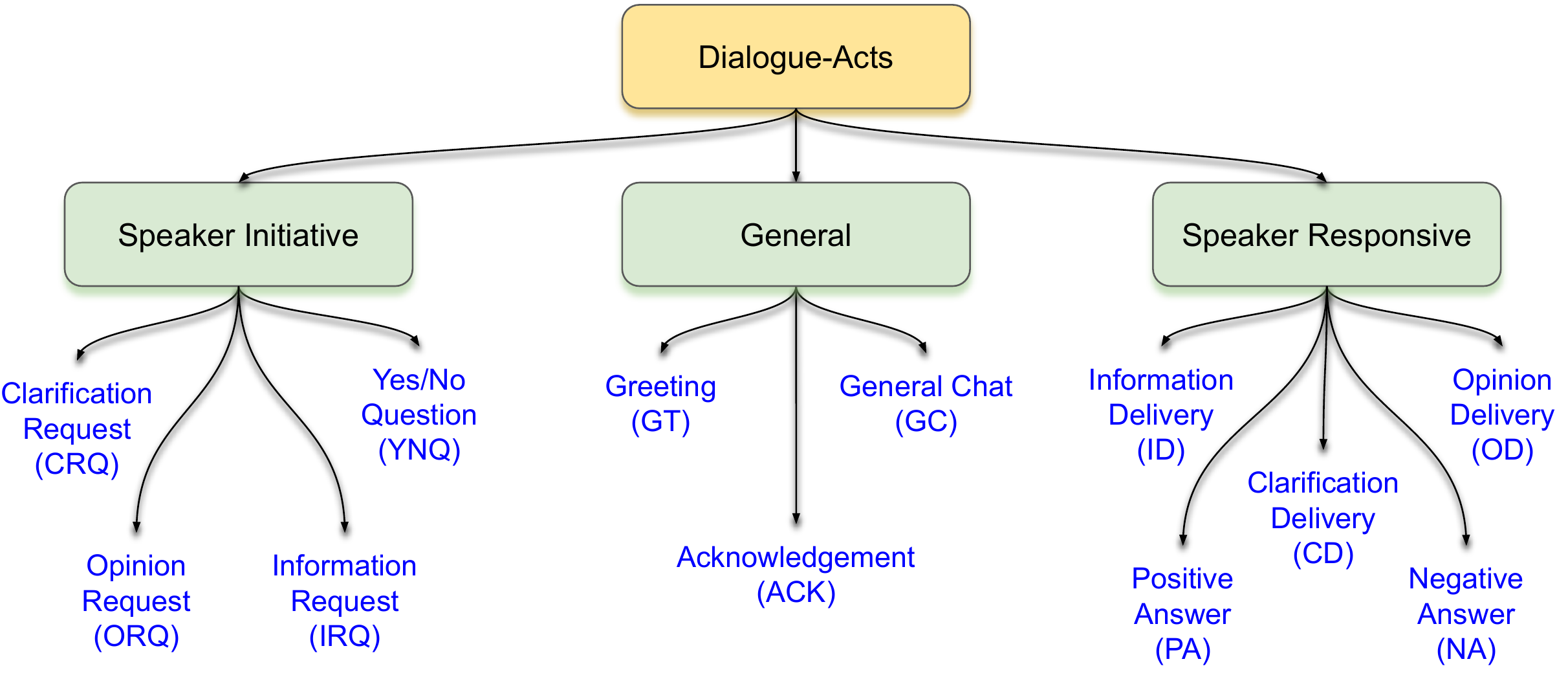}
  \vspace{-6mm}
  \caption{Annotation tree for \dataset.}
  \label{fig:DialogueAnnotationTree}
  \vspace{-5mm}
\end{figure}

\subsubsection*{\bf Dataset Annotation Scheme}
Since the counseling conversations have inherent differences with the standard conversations (such as SwitchBoard dataset conversation), it demands a carefully designed set of dialogue-act labels capable of catering the requirements of counselling conversations. Hence, we, in consultation with therapist and counselling experts, design a set of 12 dialog-act labels that are arranged in a hierarchy.
These labels are designed to capture the intents of both the patient and therapist, and also be easily comprehensible to assist in the development of a conversational dialogue system. A high-level annotation hierarchy is shown in Figure \ref{fig:DialogueAnnotationTree}. Each utterance in the dialogue belongs to one of the three categories\footnote{Sometimes, an utterance can have multiple dialog-acts; however, they are rare in the annotated dataset. Hence, for simplicity, we consider only one (primary) label for each utterance.} -- speaker initiative, speaker responsive, and general or mixed initiative. Our annotation scheme assigns three distinct dialogue-act labels to the first two categories, while the remaining four labels belong to the general category.

\begin{itemize}[leftmargin=*]
    \item \textbf{Speaker initiative labels:} When the speaker drives the conversation for the next few utterances.
        \begin{itemize}
          \item \underline{Information Request (IRQ)}: This label is used as a request for some information, e.g., `\textit{Tell me your name.}'. 
          \item \underline{Yes/No Question (YNQ)}: The YNQ label is similar to IRQ; however, the expected response is a trivial \textit{yes} or \textit{no} answer. For example, the utterance, `\textit{Did you complete your work?}' shows how a query is raised with an expected answer of {\em yes} or {\em no}. 
          \item \underline{Clarification Request (CRQ)}: This label is assigned to those utterances in which a speaker usually asks the therapist for further clarification about topic that is being currently discussed. The distinction between IRQ and CRQ is the continuation of topics -- IRQ is used whenever a discussion about a new topic or entity is started, and CRQ is used when the speaker wants to gather more information and delves deeper into the current topic at hand. For instance, if the therapist asks \textit{ You're in a situation where there is alcohol?} and follows it with another utterance  \textit{And what sort of situations are you in?}, the later utterance is an example of clarification request as the therapist delves deeper to seek causes of distress for the patient.
          \item \underline{Opinion Request (ORQ)}: The ORQ label is used when the speaker seeks opinions of the listener. For instance, the utterance \textit{How does it feel to tell me about this?} is labelled as Opinion Request.
        \end{itemize}

    \item \textbf{Speaker responsive labels:} The dialogue-act labels under this category are used in response to the speaker initiative labels. 
        \begin{itemize}
            \item \underline{Clarification Delivery (CD)}: This label is used when the speaker provides further clarifications about a topic/entity under discussion. For example, while discussing about charges that have been levied upon the patient, the therapist asks `\textit{You mean get out of the charge?}' and in response, the patient responds with the utterance \textit{Yeah. I mean, I feel like they're probably gonna make me like go to like rehab or something like that. And I don't want to do that.}. The patient response is an example of a clarification delivery. 
            
            \item \underline{Information Delivery (ID)}: When the speaker provides some factual objective information about herself. For instance, the utterances `\textit{In my last job I worked in a paper factory.}' and `\textit{I do not like my present job, I am fed up with my boss.}' are the examples of information delivery.
            
            \item \underline{Positive Answer (PA)}: These labels are used when the utterance is an answer in the form of a simple \textit{yes} to a question that was previously uttered, e.g., `\textit{Are you alright'} or `\textit{Yes}'.
            
            \item \underline{Negative Answer(NA)}:
            It is used when the utterance is an answer of the form of a simple \textit{No} to a question asked earlier.
            
            \item \underline{Opinion Delivery (OD)}: When the speaker explicitly lists out her/his opinions. For example, \textit{`You don't seem real certain.'}
        \end{itemize}
    
    \item \textbf{General:} The utterances under this category can be uttered by either of the speakers under varying circumstances. 
        \begin{itemize}
            \item \underline{Greeting (GT)}: Each session usually starts with greeting by one speaker and an appropriate response from the other, e.g., `\textit{Hello, how are you?}' and `\textit{I am fine, thank you.}'. We tag each of these utterances as GT.
            \item \underline{Acknowledgment (ACK)}: In normal conversation, very often, we utter (e.g., `\textit{Yeah! You are right.}') to acknowledge the other speaker or to show our agreement without an explicit information request, question, or command. We also observe such cases in our collected dataset; hence, we tag them as {ACK}. 
            \item \underline{General Chit-Chat (GC)}: Other utterances that do not belong to any of the above labels are tagged as GC, possibly because of the vagueness and the lack of sense in the context of the conversation. For example, the utterance \textit{'It's a beautiful day today!'} is an example of {GC}.
        \end{itemize}
\end{itemize}

\subsubsection*{\bf Annotation Process}
We employed three annotators\footnote{Annotators are in the age group of 25-35, with 2-10 years of professional experience.} who are experts in linguistics. To ensure the understanding of the tasks and annotation scheme, we took a sample of the dataset and asked each annotator to annotate them as per the prepared set of guidelines. Following this, every annotation was discussed in the presence of the annotators and an expert therapist as moderator to ensure consistency. After a couple of annotation and discussion rounds, the whole dataset was made available for the annotation. 

After the annotation process, we compute Cohen's Kappa score \cite{doi:10.1177/001316446002000104} to measure the agreement among annotators. We obtain the inter-rater agreement score of 0.7234 -- which falls under the `substantial' \cite{doi:10.1177/001316446002000104}
category.
\begin{figure}[t]
    \centering
    \includegraphics[width=0.6\columnwidth]{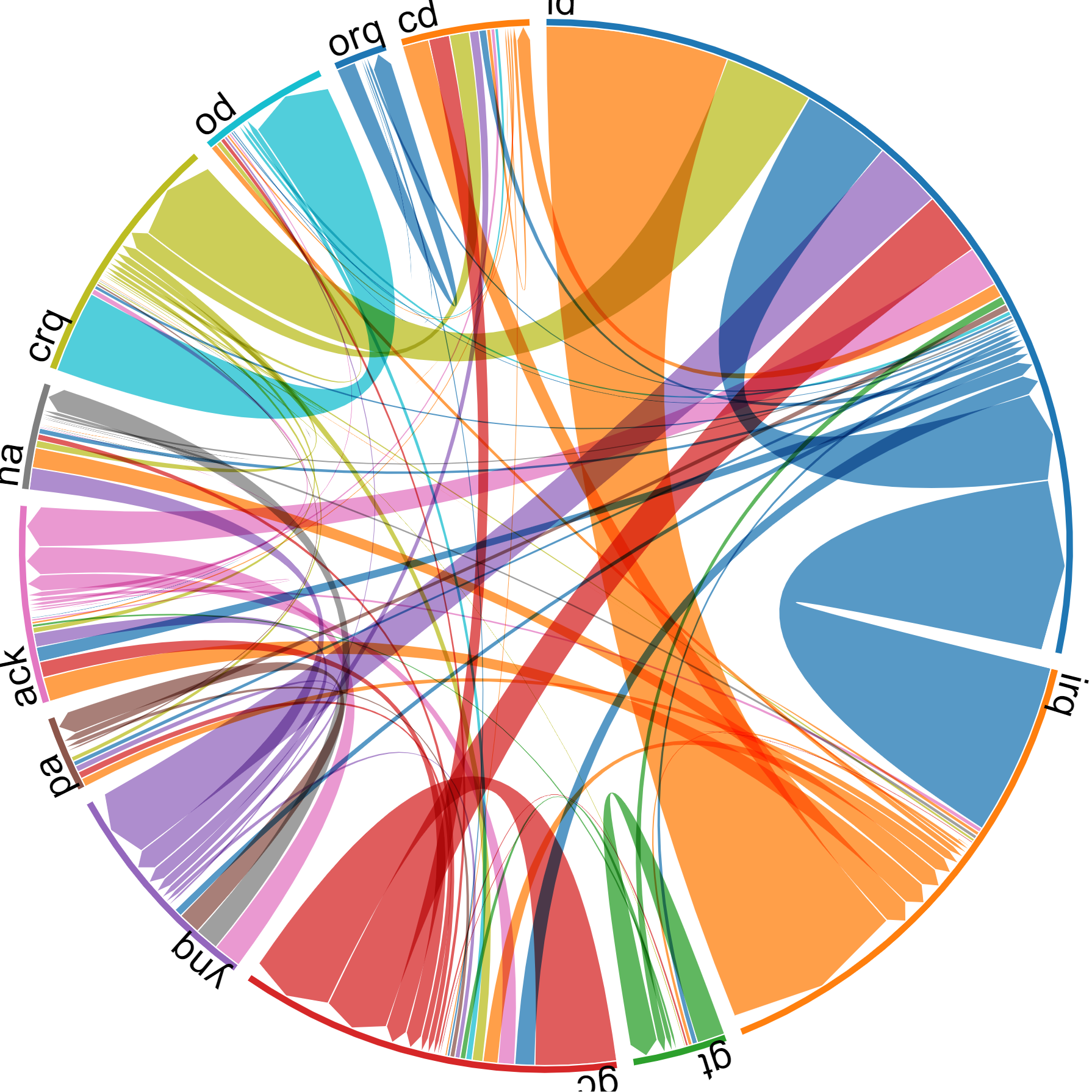}
    \caption{Relationship among the dialogue-act labels. The directed connection $U_t^x \rightarrow U_{t+1}^y$ shows the co-occurance counts of two sequential utterances with $x$ \& $y$ dialogue-acts, respectively.}
    \label{fig:chord:plot}
    \vspace{-4mm}
\end{figure}

\subsubsection*{\bf Dataset Statistics}
The broad definitions of speaker-initiative and speaker-responsive dialogue-act label pairs, IRQ \& ID; ORQ \& OD, and CRQ \& CD, seem complementary in nature. However, the dataset does not support the above view entirely. 
To show the relationship between the respective speaker-initiative and speaker-responsive dialogue-act label pairs, we present a chord diagram in Figure \ref{fig:chord:plot}. 
For any two consecutive utterances $U_t^x$ and $U_{t+1}^y$ with corresponding dialogue-act labels $x$ and $y$, each directed link ($U_t^x \rightarrow U_{t+1}^y$) between the two labels reflects their co-occurrence, and the strength of the link signifies their co-occurrence counts. Though a significant number of IRQ utterances are followed by ID utterance, in a few cases, they are followed by other dialogue-act utterances as well (e.g., ACK, GC, etc.). Similarly, ID utterances are not always preceded by IRQ utterances. We observe similar behaviour for YNQ \& PA, NA, and CRQ \& CD dialogue-act pairs as well.   
Table \ref{tab:counts} provides the statistics of the \dataset\ dataset. In total, \dataset\ has transcripts for $12.9k$ utterances
which are annotated with $12$ dialogue-act labels. These utterances are evenly distributed between the patients and therapists with $6.38K$ and $6.47K$ utterances, respectively. We split the dataset into 70:20:10 ratio as the train, test, and validation sets, respectively.

In contrast to the regular patient-doctor conversations (e.g., SOAP), the dialogue sessions in \dataset\ are usually lengthy ($\sim 59$ utterances per session). Moreover, the utterances in these sessions are themselves significantly longer as compared to other conversational datasets, with the average length of utterance of a patient being $103$ words, whereas for therapist, it is $\sim124$ words.

\begin{figure*}[ht]
  \centering
  \includegraphics[width=0.80\textwidth]{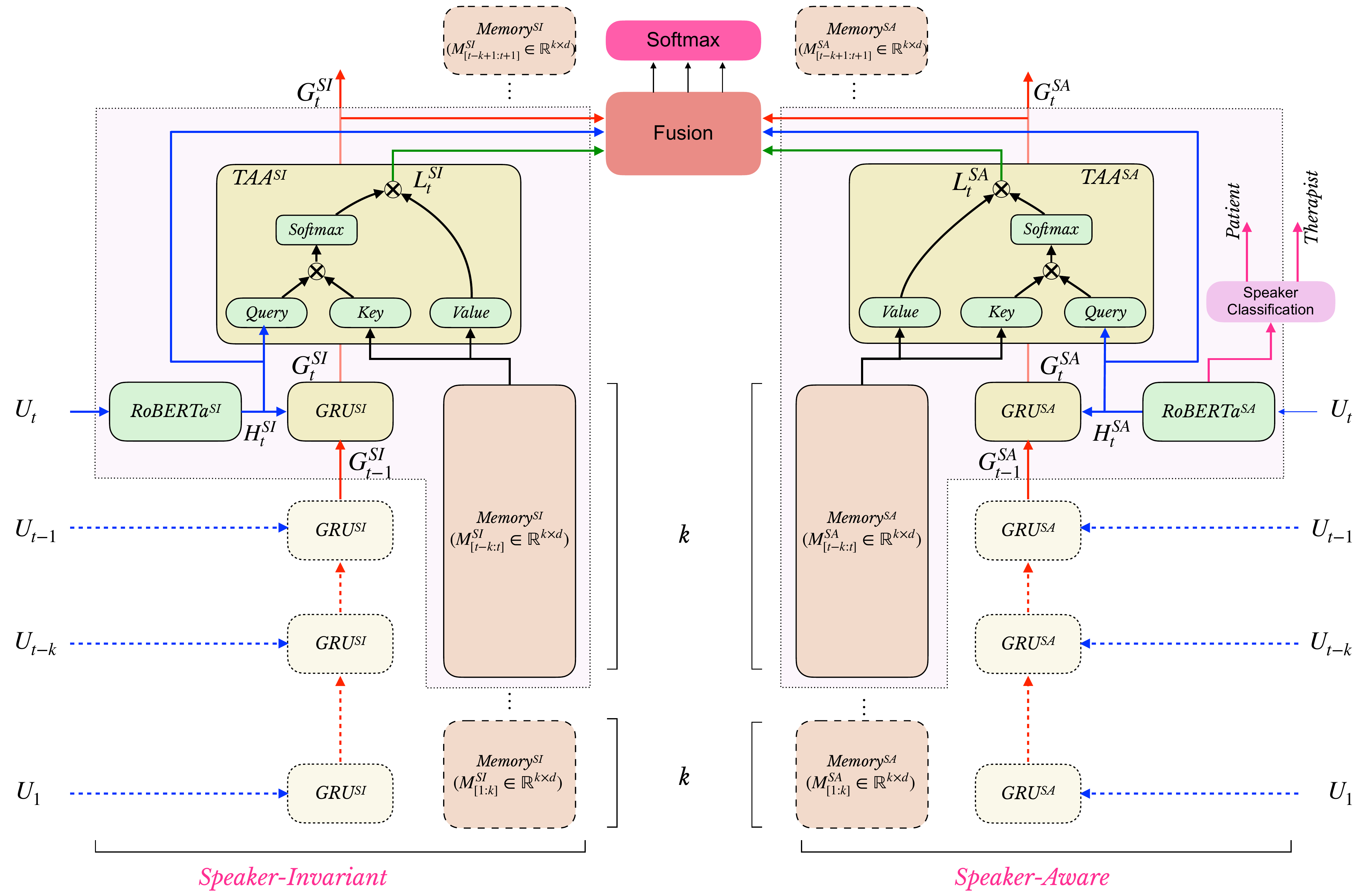}
  \vspace{-3mm}
  \caption{Architecture of \name. For each utterance $U_t$, \name\ computes the local-context $L_t$, through a time-aware attention (\textit{TAA}) mechanism on the sliding-window memory unit and the current utterance. The dialogue-level global context is maintained using a GRU $G_t$. Finally, the speaker-aware and invariant local and global contexts are fused for the task.}
  \label{fig:model-architecture}
  \vspace{-2mm}
\end{figure*}
\vspace{-6.361mm}
\section{Methodology}
We represent a therapy session as a conversation dialogue $D$ consisting of a sequence of utterances $U[1:m] = \langle U_1, U_2, ..., U_m\rangle$, where $m$ is the number of utterances in a dialogue. These utterances are uttered by the therapists and patients alternatively in a session. The objective of \name\ is to assign a correct dialogue-act label $y_t$ to every utterance $U_t$ in the dialogue.

\name\ is a transformer-based architecture that incorporates the speaker-aware contextual information for the dialogue-act classification. In our analysis of the \dataset\ dataset, we observed that a few of the dialogue-act labels are majorly associated with the patient, while a few others are related to the therapist. To model the speaker dynamics within a conversation dialogue, we consolidate a speaker-aware (SA) module in addition to the speaker-invariant (SI) module. The latter does not consider the distinction between a therapist and a patient utterance, while the former module distinguishes between the two through a pre-trained speaker identification module. Moreover, we also propose a novel time-aware attention (TAA) mechanism that considers the positions of contextual utterances during the attention computation. We hypothesize that the recent past contextual utterances have higher significance than the distant past utterances; hence, as opposed to the standard attention mechanism, TAA focuses more on the nearby (local) utterances.

For each utterance $U_t$ in a conversation dialogue, we extract the semantic representation through pre-trained RoBERTa model \cite{DBLP:journals/corr/abs-1907-11692}, which is subsequently utilized to leverage the local ($L_t$) and global ($G_t$) contextual information within the dialogue. We incorporate  a sliding-window based dynamic memory unit to compute local context through a time-aware attention mechanism. In parallel, we employ a GRU layer to capture the dialogue history as global context. 
Our analysis  reveals that a few dialogues contain utterances which discuss a topic (or an entity) that has occurred at the initial stage of the dialogue, and to correctly exploit the semantic of the utterance, the global information is desirable.
We repeat the process of local and global context extraction for both speaker-aware and speaker-invariant setups. Finally, we combine these two representations with the residual connections for the classification. Figure \ref{fig:model-architecture} shows a high-level architecture diagram of \name. 
\subsubsection*{\bf Utterance Representations}
As mentioned above, \name\ maintains two separate modules for capturing the speaker-aware and speaker-invariant information. For speaker-invariant representations, we employ a pre-trained RoBERTa language model which is further fine-tuned on DAC task. The speaker aware module is a RoBERTa model fine-tuned on Speaker Classification task. 

$H^{SI}_{t} = RoBERTa^{SI}(U^{t});\  H^{SA}_{t} = RoBERTa^{SA}(U^{t})$. 

\subsubsection*{\bf Local Context and Time-Aware Attention}
At every point in a dialogue, the nearby utterances provide important clues in the prediction of a dialogue-act label for the current utterance. For example, if the previous utterance is an information request (IRQ), then there is a good chance that the next label should be either information delivery (ID) or yes-no-answer (YNA). Therefore, we exploit the local context maintained in a memory $M_{[t-k:t]}$ for each utterance $U_t$ in the dialogue, where $k$ is the fixed local-context size. 
\begin{eqnarray}
M^{SA}_{[t-k:t]} = M^{SA}_{[(t-1)-k:(t-1)]} - M^{SA}_{[k:k+1]} + M^{SA}_{[t-1:t]} \oplus H^{SA}_t \\
M^{SI}_{[t-k:t]} = M^{SI}_{[(t-1)-k:(t-1)]} - M^{SI}_{[k:k+1]} + M^{SI}_{[t-1:t]} \oplus H^{SI}_t
\end{eqnarray}

We utilize TAA to learn the importance of contextual utterances based on their distance from the current position. At first, we pass the utterance representation ($H_t$), as computed by RoBERTa in the previous step, through a \textit{tanh} activation layer to obtain the pooler output, and subsequently project it as the \textit{query} ($q \in \mathbb{R}^{1\times d}$) vector in the attention computation. On the other hand, the contextual memory $M_{[t-k:t]}$ is projected as the \textit{key} ($K \in \mathbb{R}^{k\times d}$) and \textit{value} ($V \in \mathbb{R}^{k\times d}$) matrices. Next, we encode the local context as follows:
\begin{eqnarray}
L_t = & Softmax\left( \frac{q K^T}{D}\right) V;\ D \in \mathbb{R}^k,\ D_i = \frac{1}{i}; \  \forall k:\ 1 \le i \le k \nonumber 
\end{eqnarray}

To extract the time-aware feature, we scale-down the dot product between the \textit{query} and the \textit{key} by a monotonically decreasing function of time. The inverse function was chosen based on its empirical advantage as shown in \cite{10.1145/3097983.3097997}. The hypothesis for scaling-down the dot-product is due to the fact that as we move deeper into the dialogue history their influence on the dialogue-act reduces accordingly. Similar interaction dynamics was used in \cite{ZHANG202140}; but to our knowledge, we introduce an inverse function for the first time to compute the fixed window size attention for local context.
Following the above procedure, we compute local contexts $L_t^{SA}$ and $L_t^{SI}$ for both speaker-aware and speaker-invariant modules, respectively.
\subsubsection*{\bf Global Context} 
As the dialogue progresses, we maintain the global context of the dialogue through a GRU layer on top of the RoBERTa hidden representations. 
\subsubsection*{\bf Fusion and Final classification}
Finally, we fuse the local and global contexts of speaker-aware and speaker-invariant modules for the final classification. We also add residual connections for better gradient flow during backpropagation. Our validation results supplement the choice of concatenation as the fusion operation to be better than other operations such as global max-pooling, global mean-pooling, etc. 

\section{Experiments, Results and Analysis}
In this section, we report our experimental results, comparative study, and other analyses.
\vspace{-3mm}
\subsubsection*{\bf Baselines} 
We choose the following existing systems as baselines. 
    $\blacktriangleright$ \textbf{CASA \cite{raheja-tetreault-2019-dialogue}}: It is a context-aware attention-based system for the dialogue-act classification. It uses RNNs at dialogue and utterance levels and computes context-aware self attention before the final classification. 
    $\blacktriangleright$ \textbf{SA-CRF \cite{shang-etal-2020-speaker}:} This recent baseline incorporates a CRF layer for the classification. Moreover, it consolidates the speaker-change information using a Bi-LSTM encoder. 

In addition to these recent baselines on dialogue-act classification, we also include other sequence-labelling classification systems.
    $\blacktriangleright$ \textbf{DRNN \cite{wang-2018-disconnected}:} It is a novel Disconnected RNNs architecture which incorporates the position-invariant features for modelling. 
    $\blacktriangleright$ \textbf{ProSeqo \cite{kozareva-ravi-2019-proseqo}:} This proposed to efficiently handle the short and long texts using dynamic recurrent projections. ProSeqo avoids the use of store-and-lookup in pre-trained word-embeddings model through the context and locality-sensitive projections \cite{DBLP:journals/corr/abs-1708-00630}. 
    $\blacktriangleright$ \textbf{TextVDCNN \cite{conneau-etal-2017-deep}:} This is a deep convolutional network with residual connections for text classification. The convolutional layer works at character-level, and \textit{k}-max pooling is used to down-sample the output of convolutional layers for classification.
    $\blacktriangleright$ \textbf{TextRNN \cite{10.5555/3060832.3061023}:} This was the first work to integrate RNNs into the multi-task learning framework. We use the uniform layer architecture as described in the paper.
    $\blacktriangleright$~\textbf{RoBERTa \cite{DBLP:journals/corr/abs-1907-11692}:} We use RoBERTa as a baseline in this work due to its superioirity on various benchamarks. RoBERTa is similar to  \cite{devlin-etal-2019-bert}. 
    It is an encoder-only language model trained with masked language modelling objective on vast amounts of unlabelled data in unsupervised manner.
\subsubsection*{\bf Experimental Results}
For the experiments, we randomly split the \dataset\ dataset into $70:20:10$ ratio for the train, test, and validation sets. To measure the performances of \name\ and other baseline systems, we compute macro-F1, weighted-F1, and accuracy scores. 
We implemented our system in PyTorch \cite{NEURIPS2019_9015}
and utilized the pre-trained models from Huggingface Transformers library.

\begin{table*}[ht]
    \caption{Table showcasing the performance of baseline models as compared with our SPARTA model. $U^{t}$ represents the use of Utterance representation, GC represents Global Context, LC represents Local Context and SA means the presence of Speaker Aware representations. The dagger symbol (${\dagger}$) represents statistically significant results compared to the best baseline, CASA.}
    \label{tab:results:comparison}
    \vspace{-4mm}
    \centering
    \resizebox{0.8\textwidth}{!}
    {
    \begin{tabular}{l c c c c c c c c}
        \toprule[1pt]
        \multirow{2}{*}{\textbf{Model}} &
        \multirow{2}{*}{\textbf{Type of Modelling}} &
        \multicolumn{2}{c}{\textbf{Precision}} & 
        \multicolumn{2}{c}{\textbf{Recall}} & 
        \multicolumn{2}{c}{\textbf{F1}} & 
        \multirow{2}{*}{\textbf{Accuracy}}\\
        
         & & \bf Macro & \bf Weighted & \bf Macro & \bf Weighted & \bf Macro & \bf Weighted & \\

         \midrule[1pt]
         {TextVDCNN \cite{conneau-etal-2017-deep}} & $U^{t}$ & 11.01 & 21.02 & 19.53 & 38.53 & 13.37 & 36.81 & 41.77 \\
         {ProSeqo \cite{kozareva-ravi-2019-proseqo}} & $U^{t}$ & 9.77 & 17.60 & 11.20 & 27.90 & 7.11 & 14.29 & 27.35 \\
         {RoBERTa \cite{DBLP:journals/corr/abs-1907-11692}} &  $U^{t}$ & 51.01 & 58.12 & 47.14 & 52.97 & 43.97 & 49.13 & 52.97 \\ \cdashline{1-9}
         {TextRNN \cite{10.5555/3060832.3061023}} &  $U^{t}$~+~GC & 30.27 & 37.92 & 27.9 & 41.76 & 25.55 & 36.81 & 41.77 \\
         {DRNN \cite{wang-2018-disconnected}} &  $U^{t}$~+~GC & 28.39 & 36.72 & 31.87 & 44.32 & 28.12 & 37.82 & 44.32 \\
         {CASA \cite{raheja2019dialogue}} &  $U^{t}$~+~GC & 59.78 & 62.56 & 51.22 & 58.46 & 51.65 & 55.95 & 58.46 \\ 
         {\name-BS} & $U^{t}$ + GC & 58.94 & 62.31 & 52.02 & 57.70 & 51.83 & 54.98 & 57.70 \\ \cdashline{1-9}
         {SA-CRF \cite{shang-etal-2020-speaker}} & $U^{t}$~+~LC~+SA & 33.30 & 38.97 & 26.18 & 45.07 & 35.97 & 24.20 & 45.07 \\
         {\name-BS} & $U^{t}$ + GC + SA & 58.87 & 63.02 & 53.28 & 58.41 & 52.22 & 55.57 & 58.41 \\ 
         \midrule[1pt]
         \textbf{\name-MHA (3-fold CV)} & \multirow{2}{*}{$U^{t}$ + LC + GC + SA} & 69.60 & 71.77 & 59.45 & 62.67 & 59.00 & 62.12 & 62.67 \\
         \textbf{\name-TAA (3-fold CV)} & & 71.01 & 72.36 & 60.49 & 63.82 & 60.74 & 63.38 & 63.82 \\  \cdashline{1-9}
         
         \textbf{\name-MHA} & \multirow{2}{*}{$U^{t}$ + LC + GC + SA} & 60.24 & 66.53 & 59.64 & 63.45 & 58.16 & 63.26 & 63.45 \\
         \textbf{\name-TAA} & & 62.15 & 67.36 & 61.13 & 64.75 & 60.29$^{\dagger}$ & 64.53$^{\dagger}$ & 64.75$^{\dagger}$ \\
            
         \bottomrule[1pt]
         \multicolumn{2}{l}{Significance T-test$^{\dagger}$ (\textit{p}-value)} &  &  &  &  & 0.009 & 0.014 & 0.048 \\
         \bottomrule[1pt]
    \end{tabular}}
\end{table*}
\begin{table}[ht]
\setlength{\tabcolsep}{2pt}
\caption{Label-wise classification report for \name-TAA.}
\label{tab:classification:report}
\vspace{-3mm}
\resizebox{\columnwidth}{!}{
    \centering
    \begin{tabular}{c c c c c c c c c c c c c}
    \toprule[1pt]
        & ACK & CD & CRQ & GC & GT & ID & IRQ & OD & NA & PA & ORQ & YNQ \\
        \midrule[1pt]
        \textbf{Pre} & 44.08 & 89.91 & 50.72 & 87.38 & 60.24 & 63.76 & 70.95 & 53.19 & 58.75 & 57.33 & 48.5 & 61.36 \\
        \textbf{Rec} & 52.34 & 48.04 & 49.53 & 54.38 & 65.79 & 81.02 & 77.81 & 44.64 & 71.21 & 54.43 & 76.47 & 57.86 \\
        \textbf{F1} & 47.86 & 62.62 & 50.12 & 67.04 & 62.89 & 71.36 & 74.22 & 48.54 & 64.38 & 55.84 & 59.09 & 59.56  \\
        \toprule[1pt]
    \end{tabular}}
    \vspace{-5mm}
\end{table}

The hyperparameters are listed in Table \ref{tab-training-configuration}. The experimental results of \name\ is presented in Table \ref{tab-results-ablation} in appendix section. Since \name\ incorporates three major components -- local context, global context, and speaker-aware modules. The \name-TAA system obtains the best scores of $60.29$ macro-F1, $64.53$ weighted-F1, and $64.75\%$ accuracy, as reported in the last row of Table 2(appendix). We have also shown our observations on the ablation study on all three key-factors: Contextual information, Time-aware attention, and Speaker dynamics in ablation section in the appendix. 

We also present the label-wise performance of \name\ in Table \ref{tab:classification:report}. We can observe that \name\  consistently yields good scores for the majority of the dialogue-acts, except for the \textit{Acknowledgement} (ACK) where it records F1-score of merely 47.86\%. Even for the under-represented labels ({ORQ},{NA} and {PA}) in \dataset, \name\ reports good F1-scores of 59.09\%, 64.38\% and 55.84\%, respectively for these three labels. 

\subsubsection*{\bf Comparative Analysis:}
We compare \name\ with various existing systems and other baselines. The comparative analysis is reported in Table \ref{tab:results:comparison}. Based on the type of modelling, we categorize the baselines into three groups -- \textit{utterance-driven} ($U_t$), \textit{utterance+global context driven} ($U_t$ + GC), and \textit{utterance + global context + speaker-aware driven} ($U_t$ + GC + SA). Comparatively, \name\ incorporates local context in addition to utterance, global context, and speaker dynamics ($U_t$ + LC + GC + SA). In the first category, the standard \textit{RoBERTa} model attains the best macro-F1, weighted-F1 and accuracy of $43.97$, $49.13$, and $52.97\%$, respectively. In comparison, CASA \cite{raheja-tetreault-2019-dialogue} yields the improved weighted-F1 and accuracy scores at $55.95\%$ ($+6.82\%$) and $58.46\%$ ($+5.49\%$), respectively, with the global context as an additional information. Finally, we experiment with SA-CRF \cite{shang-etal-2020-speaker} which also includes the speaker-dynamics for the dialogue-act classification; however, its performance on \dataset\ is not at par with CASA \cite{raheja-tetreault-2019-dialogue}. It reports $35.97$, $24.20$, and $45.07\%$ macro-F1, weighted-F1, and accuracy, respectively.
In comparison, \name-TAA\ obtains significant improvements over all baselines. It reports improvements of $+8.64\%$, $+8.58\%$, and $+6.29\%$ in macro-F1 ($60.29$), weighted-F1 ($64.53$), and accuracy ($64.75\%$), respectively, as compared to CASA suggesting the incorporation of local context extremely effective. Note that ProSeqo \cite{kozareva-ravi-2019-proseqo} and CASA \cite{raheja-tetreault-2019-dialogue} are currently the state-of-the-art on switchboard dialogue-act corpus benchmark\footnote{https://paperswithcode.com/sota/dialogue-act-classification-on-switchboard}; yet they report inferior scores on \dataset\ compared to \name. Moreover, we also report the mean of the 3-fold cross-validation results for both \name-MHA and \name-TAA, and the results are consistent with the train-val-test split case. We also perform statistical significance T-test comparing \name-TAA and the best performing baseline (CASA). We observe that our results are significant with $>95\%$ confidence across macro-F1 (\textit{p}-value$=0.009$), weighted-F1 (\textit{p}-value$=0.014$), and accuracy (\textit{p}-value$=0.048$) values.
\begin{figure}[ht]
    {\includegraphics[width=0.37\textwidth]{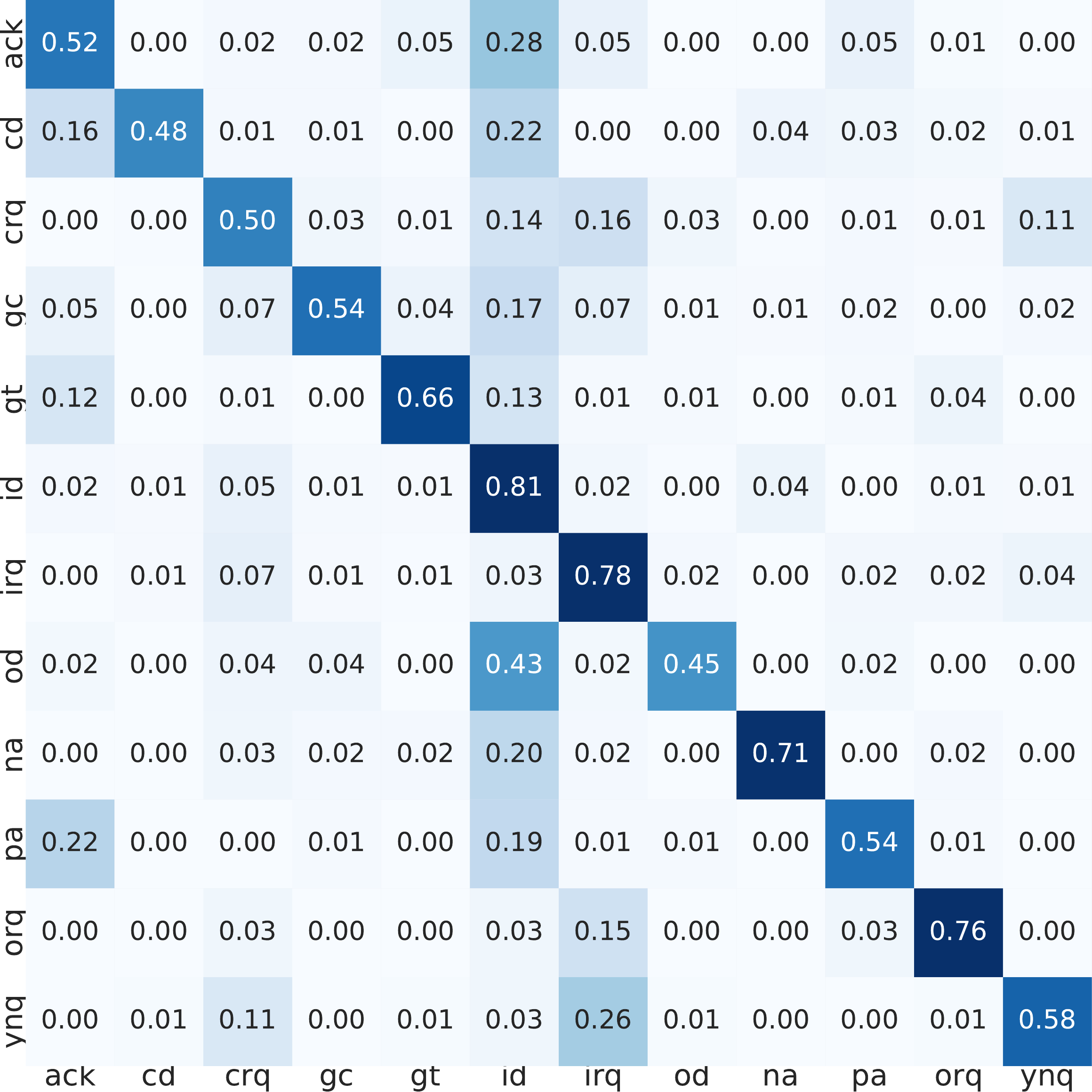}} 
    \vspace{-2mm}
    \caption{Confusion matrix for \name-TAA.}
    \label{fig:confusion-matrix}
    \vspace{-6mm}
\end{figure}

\begin{table*}[t]
    \caption{An example therapy session showcasing the differences in predictions between the baseline and \name. We truncate the dialogue due to its length (total 151 utterances).}
    \label{tab:preds}
    \vspace{-4mm}
    \centering
    \resizebox{0.92\textwidth}{!}
    {
    \begin{tabular}{c p{35em} c c c c}       
    \toprule[1pt]
        & \multirow{2}{*}{\bf Utterances} & \multirow{2}{*}{\bf Ground Truth} & \multicolumn{3}{c}{\bf Prediction} \\ \cline{4-6}

        &  & & CASA & \name-TAA & \name-MHA \\ \toprule[1pt]
         & \textcolor{blue}{Therapist:} \textit{Hi Angela, how are you doing today ?} & GT & GT & GT & GT\\
         & \textcolor{red}{Patient:} \textit{I am all right, thanks} & GT & GT & GT & GT\\
         & ... \textit{....} & .. & .. & .. & ..\\
         & \textcolor{blue}{Therapist:} \textit{How's your mood overall?} & IRQ & IRQ & IRQ & IRQ\\
         & \textcolor{red}{Patient:} \textit{\colorbox{olive!30}{I mean, what can I say I lost my job. I'm seeing you because I have a drinking} \colorbox{olive!30}{problem. So things are not that great.}} & ID & ID & ID & \colorbox{red!30}{CD}\\
         & ... \textit{....} & .. & .. & .. & ..\\
         & \textcolor{blue}{Therapist:} \textit{When you're drinking does it still feel good?} & IRQ & IRQ & IRQ & IRQ\\
         & \textcolor{red}{Patient:} \textit{Nothing really feels good.} & ID & ID & ID & ID\\
         & \textcolor{blue}{Therapist:} \textit{Just gets back to normal?} & CRQ & CRQ & CRQ & CRQ\\
         & \textcolor{red}{Patient:} \textit{Yeah, pretty much.} & CD & \colorbox{red!30}{ID} & CD & \colorbox{red!30}{ID}\\
         & \textcolor{blue}{Therapist:} \textit{When was the last time you drank ?} & IRQ & IRQ & IRQ & IRQ\\
         & \textcolor{red}{Patient:} \textit{This morning.} & ID & ID & ID & \colorbox{red!30}{GC}\\ 
         & ... \textit{....} & .. & .. & .. & ..\\
         & \textcolor{red}{Patient:} \textit{\colorbox{olive!30}{Oh, I should stop. I mean, like, I'm not, not unrealistic.}} & OD & \colorbox{red!30}{ID} & OD & \colorbox{red!30}{ID}\\
         & \textcolor{blue}{Therapist:} \textit{Yeah} & ACK & \colorbox{red!30}{OD} & ACK & ACK \\ \toprule[1pt]
    \end{tabular}
    }
    \vspace{-4mm}
\end{table*}
\subsubsection*{\bf Error Analysis: }
In this section, we present two-way error analyses of \name\ in terms of quantitative and qualitative evaluations. 
\noindent\textit{Quantitative analysis:} We report the confusion matrix for \name-TAA in Figure \ref{fig:confusion-matrix}. We observe three pairs with significant error-rates ($\ge 25\%$) -- YNQ:IRQ ($26\%$), OD:ID ($43\%$), ID:ACK ($28\%$). For the prediction of \textit{information delivery} (ID), \name\ is confused most of the time with other classes -- $19\%$ with PA, $20\%$ with NA, $43\%$ with OD, $13\%$ with GT, $17\%$ with GC, $14\%$ with CRQ, $28\%$ with ACK, and $22\%$ with CD.. We can relate this behaviour to the diversity of the utterances with ID tag, i.e., the discussion in these utterances generally contains a fair segments of utterances from other dialogue-acts (e.g., `\textit{Yeah, That's something I always do.}' could be easily be confused with PA). The other prominent error-case is found in IRQ:YNQ pair, we observe a confusion of $26\%$ between IRQ and YNQ utterances because of the versatile questioning behaviour. For the remaining cases, error-rates are nominal. Thus, we articulate that \name\ can be further improved with more balanced dataset.          

\noindent \textit{Qualitative analysis: } Table \ref{tab:preds} shows a sample session along with actual and the predicted dialogue-act labels for \name\ and the best baseline model (i.e., CASA). Due to the length of the conversation, we truncate some of the utterances in between; however, the gist of the conversation is that {\em the patient is stressed of losing her job and having drinking issues and the therapist is trying to understand the core problem}. The conversation has mostly \textit{information request} and \textit{information delivery} types of utterances with a few other dialogue-acts in between (e.g., \textit{CRQ} \textit{GT}, etc.). 

We can observe that for the first three utterances, \name\ and the baseline are consistent with the actual labels. For the fourth utterance, the \name-MHA model misclassifies the utterance as CD when the patient is clearly providing objective information which is necessary for further conversation. In the seventh utterance, the therapist wants more clarification about the drinking habits, and the patient provides the clarification but CASA and \name-TAA models wrongly classify this utterance as ID when there is no objective information being provided here. Next, we notice that when the patient talks about stopping this and provides her opinion that she is \textit{not unrealistic}, the \name-MHA and CASA models predict the wrong label for this utterance as ID. In the next utterance, the therapist acknowledges this opinion, but the CASA model predicts the wrong label. So, not only the \name-TAA model is able to capture the semantics of the utterances better, it also utilizes the contextual information in a better way by relating the past information about the speaker with the current utterances.

\noindent {\bf General Discussion: }
The work presented in this paper was motivated solely by the dire need of understanding conversations that occur in counselling sessions and to design solutions that would help the therapists to better understand the intents of their patients. However, the proposed model can be easily adaptable to other domain (such as normal chit-chat-based conversation) as well. To ensure we do not deviate from the prime objective of this work, we restrict ourselves to explore the dialog-act classification in the counselling conversations only.

\vspace{-0.3cm}
\section{Limitations and Ethical Considerations}
\noindent
We aim to tackle a very sensitive and a pervasive public-health crisis. We transcribe the data from publicly available counselling videos. The automatic transfer of utterance from the speech modality to the text causes some information loss, though we tried our best to recover them through manual intervention. Moreover, we consulted with mental-health professionals and linguists in preparing the annotation guideline. However, the annotator's bias cannot be ruled out completely. The names of the patients and therapists involved in these sessions have been systematically masked.
Another important aspect of the current work is that the majority of the sessions in \dataset\ belong to the mental health professionals and patients based in United States. Hence, the effectiveness of \name\ on data from other geographical or demographical regions may vary. 
We understand that the building computational models in mental-health avenues has high-stakes associated with it and ethical considerations, therefore, become necessary. No technology will work perfectly in solving the problems related to mental health \cite{Miner2020AssessingTA}. It is important to note that we do not make any diagnostic claims. Further, the deployment of any such technology will be done keeping in mind the safety-risks and mitigating any sources of bias that may arise.
\vspace{-5mm}
\section{Conclusion and Future Work}
Paying heed to the consequences of the COVID-19 pandemic on mental health, in this paper, we raised the attention on the much deserved research on dialogue system for mental-health counselling. To this end, we collected and developed the \dataset\ dataset for the dialogue-act classification in dyadic counselling conversations. We defined twelve  dialogue-act labels to cater to the requirement of counselling sessions. In total, we annotated $\sim 12.9k$ utterances across $212$ sessions. We also proposed \name, a novel transformer-based speaker and time-aware joint contextual learning model for dialogue-act classification.
\name\ utilizes the global and local context in the speaker-aware and speaker-invariant setups while also using a novel memory-driven time-aware attention mechanism to leverage the local context.
Our extensive ablation study and the comparative analysis established the superiority of \name\ over several existing models.
In future, we would like to extend our effort in the development of dialogue-systems for mental-health counselling by including other crucial tasks such as emotion recognition, dialogue summary generation, dialogue-state tracking, empathetic response generation, etc.
Another dimension of the future work is to include other languages and demographic diversities to cater the requirements of a larger population.


\bibliographystyle{ACM-Reference-Format}
\bibliography{references}
\appendix
\appendixpage
\addappheadtotoc


\section{Speaker Aware and Speaker Invariant Representations}
\begin{figure}[hbt!]
  \centering
  \includegraphics[width=0.20\textwidth]{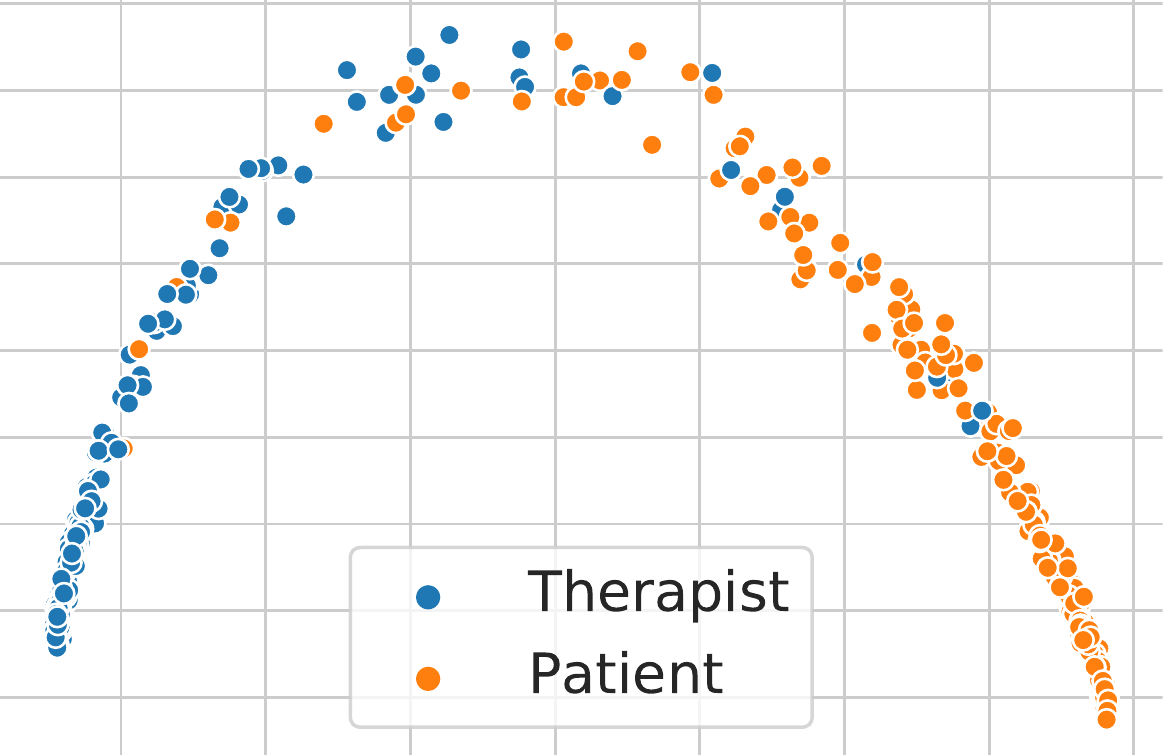}
  \hspace{6mm}
  \includegraphics[width=0.20\textwidth]{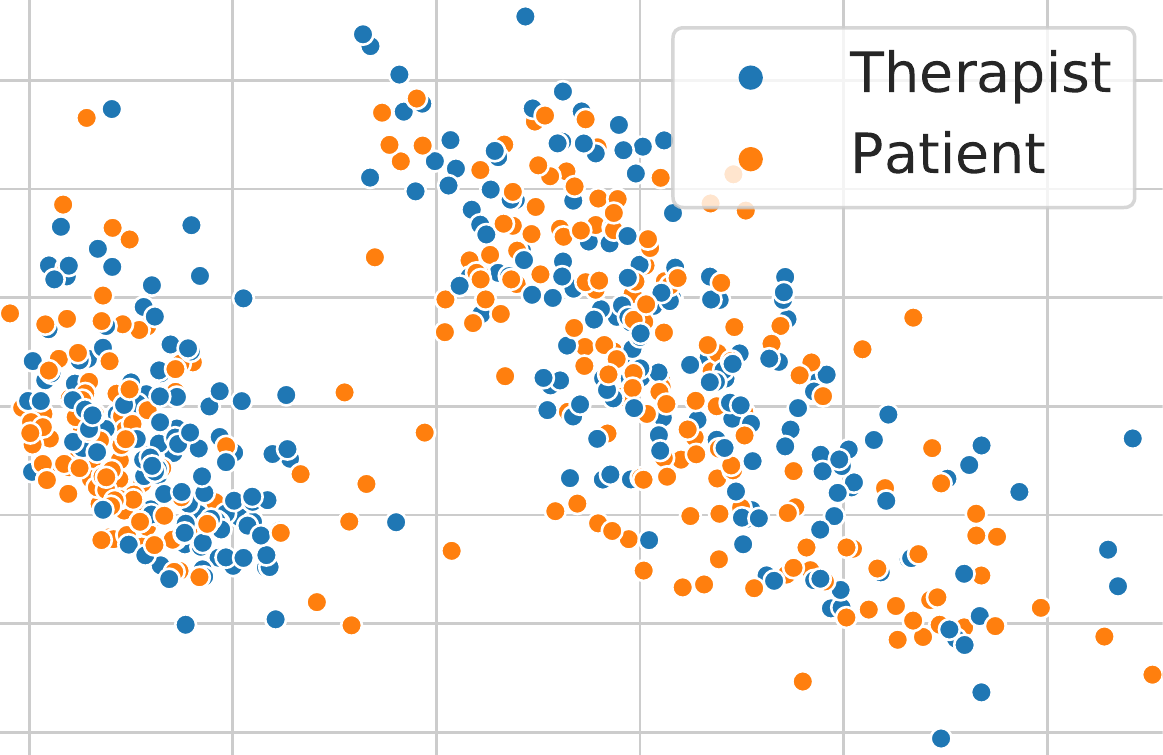}
    \caption{Speaker-aware (left) and speaker-invariant (right) utterance representations for the two speakers in a 2D space.}
  \label{fig:utterance-representation}
  \vspace{-4mm}
  
\end{figure}

Furthermore, in the quest of explanation, we visualize the speaker-aware ($H_t^{SA}$) and speaker-invariant ($H_t^{SA}$) utterance representation in Figure \ref{fig:utterance-representation}. We employ principal component analysis   to project the representations into 2D. We can observe that the speaker-aware representation distinguishes between the patient's and therapist's utterances pretty well. In contrast, the patient and therapist utterances are mixed in the speaker-invariant representation. Considering the underlying task, where the therapist and the patient have the major contributions in speaker-initiative and speaker-responsive dialogue-acts, speaker-aware representation provides crucial assistance to \name.  

\section{Reproducibility}
We use PyTorch \footnote{https://pytorch.org/} and Pytorch Lightning \footnote{https://www.pytorchlightning.ai/} frameworks for all our experiments. We extensively used Hugging Face Library for implementation of transformer based NLP models.

\subsubsection{Hardware Utilization: } The complete model training was done on a linux machine with following specifications:
\begin{itemize}
    \item GPU: TeslaV100
    \item Memory: 32GB
    \item Linux 64 Bit: Ubuntu 18.04 LTS
\end{itemize}

\subsubsection{Output Logs: }We use WandB\footnote{https://wandb.ai/} library to log all our results.

\subsubsection{Experiment Details}
The \texttt{categorical cross-entropy} loss is optimized using the \texttt{Adam} optimizer. We use \texttt{dropout}$=0.15$ and train the models for 50 epochs in a batch of 8 samples. We use EarlyStopping to prevent our model from overfiting. A comprehensive list of hyperparameters is given in \ref{tab-training-configuration}.

We use pretrained RoBERTa 
available from the Hugginface\footnote{https://huggingface.co/} transformers
for speaker aware as well as speaker invariant utterance representations. We create separate classes for each sub-module in our model. Built-in dataset and dataloader classes are used to load the data in the correct format and a total of 6 workers are used to load the data. 
For the GRU, we use input size and hidden dimension of 768 with 1 layer. For the classifier sub-module, we use a dropout of 0.1 a linear layer of 768 neurons and LeakyReLU as the activation. Our attention blocks comprise of 3 Linear Layers for keys,queries and values and one linear layer to project the keys, queries and values.
\subsubsection{Requirements}
\begin{itemize}
    \item pytorch==1.7.1
    \item pytorch-lightning==1.10
    \item transformers==4.1.0
    \item scikit-learn==0.24.0
    \item numpy==1.19.4
    \item pandas==1.1.5
    \item wandb==0.10.12
    \item datasets==1.1.3
\end{itemize}

\begin{table}[t]
    \caption{Training hyperparameters}
    \begin{tabular}{l c}
        \toprule[1pt]
         \textbf{Hyperparameter} & \textbf{Value} \\
         \midrule[1pt]
         Max Length & 512 \\
         {Batch Size} & 8 \\
         {Hidden Size} & 768 \\
         {Layers}\textsubscript{\textsubscript{GRU}} & 1 \\
         {Layers}\textsubscript{\textsubscript{RoBERTa}} & 12 \\
         {Dropout} & 0.15 \\
         Window Size & 6 \\
         Learning Rate & 1e-5 \\
         Precision & 32 \\
         Min Delta & 0.001 \\
         Patience & 5 \\
         Epochs & 50 \\
         \hline
    \end{tabular}
    \label{tab-training-configuration}
\end{table}

\begin{table*}[hbt!]
    \centering
    \caption{Experimental results of \name\ on \dataset\ Dialogue-act Corpus. TAA: \textit{Time-aware attention}; MHA: \textit{Multihead attention}; BS: \textit{Baseline - without attention}; LC: \textit{Local context}; GC: \textit{Global context}; SA: \textit{Speaker-aware}. Best results are highlighted in bold.}
    
    \resizebox{0.8\textwidth}{!}
    {
    \begin{tabular}{ccccccccccc}
    \toprule[1pt]
        \multirow{2}{*}{\bf Model} & \multirow{2}{*}{\bf LC}& \multirow{2}{*}{\bf GC}&
        \multirow{2}{*}{\bf SA}&
        \multicolumn{2}{c}{\bf Precision} & 
        \multicolumn{2}{c}{\bf Recall} & 
        \multicolumn{2}{c}{\bf F1} & 
        \bf Accuracy \\
         & & & & \bf Macro & \bf Weighted & \bf Macro & \bf Weighted & \bf Macro & \bf Weighted & \\
        \toprule[1pt]
        \multirow{2}{*}{\name-BS} & - & \checkmark & - & 58.94 & 62.31 & 52.02 & 57.70 & 51.83 & 54.98 & 57.70 \\
        & - & \checkmark & \checkmark & 58.87 & 63.02 & 53.28 & 58.41 & 52.22 & 55.57 & 58.41 \\ \cdashline{1-11}
        \multirow{3}{*}{\name-MHA}  & \checkmark & - & - & 51.57 & 61.11 & 50.01 & 53.10 & 46.51 & 53.25 & 53.10 \\
        & \checkmark & \checkmark & - & 58.99 & 63.06 & 53.00 & 58.23 & 52.24 & 55.49 & 58.23 \\
        & \checkmark  & \checkmark & \checkmark & 60.24 & 66.53 & 59.64 & 63.45 & 58.16 & 63.26 & 63.45 \\ \cdashline{1-11}
        \multirow{3}{*}{\name-TAA}  & \checkmark & - & - & 52.90 & 60.29 & 49.87 & 54.84 & 47.63 & 53.72 & 54.84 \\
        & \checkmark & \checkmark & - & 60.25 & 66.25 & 57.23 & 59.08 & 53.73 & 55.07 & 59.08 \\
        & \checkmark  & \checkmark & \checkmark &  62.15 & 67.36 & 61.13 & 64.75 & 60.29 & 64.53 & 64.75\\
    \bottomrule
    \end{tabular}}
    \label{tab-results-ablation}
\end{table*}

\subsubsection{Training and Evaluation}
\begin{itemize}
    \item Create a python environment and install the above libraries.
    \item Prepare the data and store it in csv files and change the data directory in the config.py file in \textsc{'data\_dir'} and \textsc{'data\_files'} variables.
    \item The \texttt{models/} directory contains the code of all the sub-modules.
    \begin{itemize}
        \item The file \texttt{Classifier.py} contains the code of a generic classifier that classifies data into two classes.
        \item \texttt{SpeakerClassifier.py} contains the code of the speaker classifier. It has been isolated so that ablation studies can be performed easily.
        \item \texttt{MHA.py} contains the code of our Multi-Head attention module whereas \texttt{Relevance.py} contains the code of our Time Aware Attention module.
        \item \texttt{Roberta.py} and \texttt{GRU.py} contains the code to initialize RoBERTa and GRU models respectively.
        \item \texttt{DAC.py} contains the code of our complete architecture.
    \end{itemize}
    \item The file \texttt{dataset.py} in \texttt{dataset/} directory contains the code to create a custom dataset class and load data using the torch.utils.data.Dataset and torch.utils.data.DataLoader wrappers.
    \item Execution
    \begin{itemize}
        \item \texttt{Trainer.py} contains the code to create pytorch-lightning trainer with \texttt{wandb} logger to log the results.
        \item Use \texttt{config.py} to change the configuration of models or to change the hyperparameters used in the experiments.
        \item Run \texttt{python main.py} to run the code and get results.
    \end{itemize}
\end{itemize}

\section{Ablation Study}
\label{ablation-section}
To verify the effectiveness of the time-aware attention to compute the local context, we also report the results with standard multi-head attention (MHA) 
mechanism. 
The first set of results (i.e., \name-BS or the baseline version of \name) in Table \ref{tab-results-ablation} represents the absence of local context in \name. It yields macro and weighted F1-scores of $51.83$ and $54.98$ respectively with global context only. We obtain $57.70\%$ accuracy for the same setup. On the other hand, incorporating the speaker-aware information into the system improves the trio of accuracy, macro, and weighted F1-scores by $[1-2]$ points, suggesting a positive impact of speaker-dynamics in \name. Subsequently, we introduce the local context with multi-head attention in \name-MHA. In comparison with \name-BS, \name-MHA reports $52.24$ macro-F1 and $55.49\%$ weighted-F1 with both the local and global contexts. Furthermore, we obtain $58.16$  $(+5.92\%)$ and $63.26$ $(+7.77\%)$ macro and weighted F1-scores, respectively, with the inclusion of speaker-aware module in the system as compared to \name-MHA without the Speaker-Aware module. Moreover, it yields better accuracy at $63.45\%$ $(+5.22\%)$. These results clearly indicate the importance of local context and speaker-aware modules for the dialogue-act classification. In the final stage of our experiments, we replace the multi-headed attention with the proposed time-aware attention mechanism (TAA). 

\subsection{Observations from Ablation Study}
We summarize our observation as follows:
\begin{itemize}[leftmargin=*]
    \item \textbf{Contextual information:} Dialogue-act labels in a counseling based conversation not only depend on the abstract information of the dialogue but also on the utterances in the immediate vicinity of the current utterance. As can be observed from Table \ref{tab-results-ablation}, the presence of both the global and local contextual information plays an importance role in \name. Moreover, the absence of either of the component degrades the overall performance. This corroborates that they carry distinct and diverse contextual information.
    \item \textbf{Time-aware attention:} The comparison between the standard multi-head attention and the novel time-aware attention highlights the importance of attending over the relative positions of utterances. As stated earlier, extensive experimental results show that TAA yields better performance compare to MHA for all possible configurations.
    \item \textbf{Speakers dynamics:} For all combinations, we observe performance drop of $[3-4]\%$ without the speaker information. This is particularly apparent as a majority portion of the counseling conversation is driven by the therapist, thus the speaker-initiative dialogue-acts have higher relevance with the therapist utterances. Therefore, our intuition of incorporating the speaker-aware module as a critical component in \name\ also corroborates the empirical evidences. 
\end{itemize}
\section{Role of Emotion}
In earlier phases of experiments, we also considered the role of emotion in dialogue act classification. We annotated a subset of our dataset with three emotion classes consisting of `positive', `negative' and `neutral'. We found that around $\sim70\%$ of utterances by the patients belonged to `negative' class whereas $\sim90\%$ of therapists' utterances belonged to `neutral' class. Due to such imbalance, this data was not used in the final version of our proposed architecture.

\end{document}